\documentclass[10pt,twocolumn,letterpaper]{article}

\usepackage{cvpr}              % To produce the CAMERA-READY version
\usepackage{graphicx}
\usepackage{amsmath}
\usepackage{amssymb}
\usepackage{booktabs}
\usepackage{mathtools}

\makeatletter
\@namedef{ver@everyshi.sty}{}
\makeatother
\usepackage{textcomp}
\usepackage{pgfplots}
\pgfplotsset{width=10cm, compat=1.9}

\definecolor{color1}{RGB}{230, 159, 0}
\definecolor{color2}{RGB}{86, 180, 233}
\definecolor{color3}{RGB}{0, 158, 115}
\definecolor{color4}{RGB}{240, 228, 66}
\definecolor{color5}{RGB}{0, 114, 178}
\definecolor{color6}{RGB}{213, 94, 0}
\definecolor{color7}{RGB}{204, 121, 167}

\pgfplotscreateplotcyclelist{colorblind-friendly}{
{color1!60!black,fill=color1},
{color2!60!black,fill=color2},
{color3!60!black,fill=color3},
{color4!60!black,fill=color4},
{color5!60!black,fill=color5},
{color6!60!black,fill=color6},
{color7!60!black,fill=color7},
}

\usepackage[pagebackref,breaklinks,colorlinks]{hyperref}
\usepackage{caption}

\usepackage[capitalize]{cleveref}
\crefname{section}{Sec.}{Secs.}
\Crefname{section}{Section}{Sections}
\Crefname{table}{Table}{Tables}
\crefname{table}{Tab.}{Tabs.}

\usepackage{enumitem}
\setlist{topsep=2pt, parsep=2pt, leftmargin=*}

\def\cvprPaperID{11298} % *** Enter the CVPR Paper ID here
\def\confName{CVPR}
\def\confYear{2022}

\usepackage{overpic}
\usepackage{enumitem} %< control spacing in itemize/enumerate/...
\usepackage{overpic} %< add raw math symbols to figures
\usepackage{color}
\usepackage{comment}
\usepackage{gensymb}
\definecolor{turquoise}{cmyk}{0.65,0,0.1,0.3}
\definecolor{purple}{rgb}{0.65,0,0.65}
\definecolor{dark_green}{rgb}{0, 0.5, 0}
\definecolor{orange}{rgb}{0.8, 0.6, 0.2}
\definecolor{red}{rgb}{0.8, 0.2, 0.2}
\definecolor{darkred}{rgb}{0.6, 0.1, 0.05}
\definecolor{blueish}{rgb}{0.0, 0.3, .6}
\definecolor{light_gray}{rgb}{0.7, 0.7, .7}
\definecolor{pink}{rgb}{1, 0, 1}
\definecolor{greyblue}{rgb}{0.25, 0.25, 1}

\newcommand{\Ours}{AutoRF\xspace}

\usepackage{blindtext}

\renewcommand{\paragraph}[1]{\vspace{1em}\noindent\textbf{#1}.}
\begin{document}

\newcommand{\myparagraph}[1]{\vspace{3pt}\noindent\textbf{#1.}}

%\title{Learning 3D Object Priors with Implicit Autoencoders}
%\title{Learning 3D Object Priors from Single View Observations}
\title{\vspace{-40pt}AutoRF: Learning 3D Object Radiance Fields from Single View Observations}

\author{
Norman M{\"u}ller$^{1,3}$~~~
Andrea Simonelli$^{2,3}$~~~ 
Lorenzo Porzi$^3$~~~
Samuel Rota Bul\`{o}$^3$~~~\\
Matthias Nie{\ss}ner$^1$~~~
Peter Kontschieder$^3$~~~
\vspace{0.2cm} \\
Technical University of Munich$^1$~~~
University of Trento$^2$~~~
Meta Reality Labs Zurich$^3$
\vspace{0.2cm}
% \and
}

\begin{comment}
\author{Norman M{\"u}ller*\\
Technical University of Munich\\
\and
Andrea Simonelli*\\
University of Trento\\
\and
Lorenzo Porzi\\
Meta Reality Labs\\
\and
Samuel Rota Bul\`{o}\\
Meta Reality Labs\\
\and
Matthias Nie{\ss}ner\\
Technical University of Munich
\and
Peter Kontschieder\\
Meta Reality Labs\\
}
\end{comment}

\twocolumn[{%
	\renewcommand\twocolumn[1][]{#1}%
	\begin{center}
\maketitle
        \vspace{-16pt}
		\includegraphics[width=1.0\linewidth]{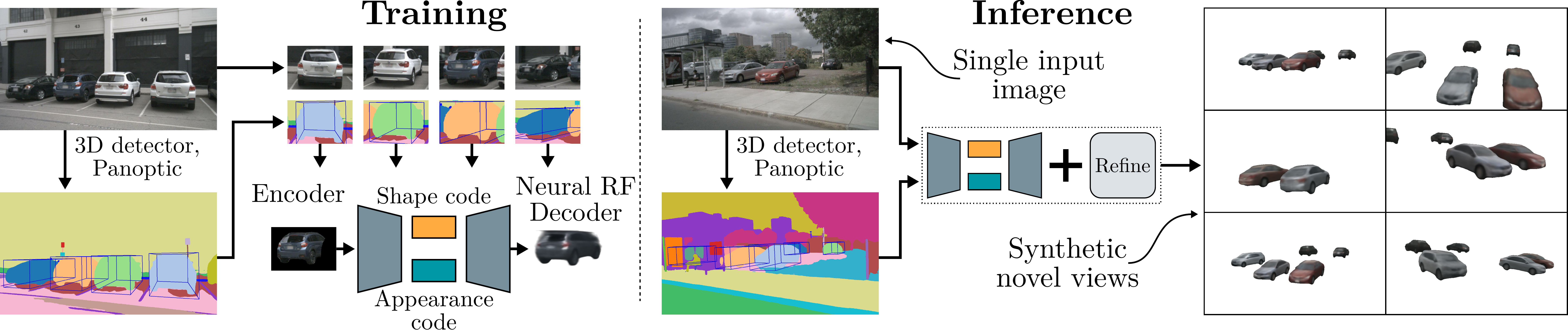}
	\end{center}    
	\vspace{-10pt}
	\captionof{figure}{Overview of AutoRF. Our model consists of an encoder that extracts a shape and an appearance code from an object's image, which can be decoded into an implicit radiance field operating in normalized object space and leveraged for novel view synthesis. Object images are generated from real-world imagery by leveraging machine-generated 3D object detections and panoptic segmentation. At test time, we fit objects to respective target instances using a photo-metric loss formulation.}
	\label{fig:teaser}
	\vspace{0.3cm}
}]

\newcommand\blfootnote[1]{%
  \begingroup
  \renewcommand\thefootnote{}\footnote{#1}%
  \addtocounter{footnote}{-1}%
  \endgroup
}

\begin{abstract}
\vspace{-8pt}
We introduce \Ours~-- a new approach for learning neural 3D object representations where each object in the training set is observed by only a single view.
This setting is in stark contrast to the majority of existing works that leverage multiple views of the same object, employ explicit priors during training, or require pixel-perfect annotations.
To address this challenging setting, we propose to learn a normalized, object-centric representation whose embedding describes and disentangles shape, appearance, and pose. 
Each encoding provides well-generalizable, compact information about the object of interest, which is decoded in a single-shot into a new target view, thus enabling novel view synthesis.
We further improve the reconstruction quality by optimizing shape and appearance codes at test time by fitting the representation tightly to the input image.
In a series of experiments, we show that our method generalizes well to unseen objects, even across different datasets of challenging real-world street scenes such as nuScenes, KITTI, and Mapillary Metropolis. Additional results can be found on our project page \url{https://sirwyver.github.io/AutoRF/}.
\blfootnote{Work was done during Norman's and Andrea's internships at Meta Reality Labs Zurich.}
\end{abstract}
\vspace{-15pt}
\section{Introduction}
In this work, we address the challenging problem of inferring 3D object information from individual images taken in the wild. Providing an objects’ 3D shape with 6DOF pose and corresponding appearance from a single image is key to enabling immersive experiences in AR/VR, or in robotics to decompose a scene into relevant objects for subsequent interaction. The underlying research problem is related to novel view synthesis or inverse graphics, and has recently gained a lot of attraction in our community \cite{Henderson_2020_CVPR,cmrKanazawa18,saito2019pifu,Wu_2020_CVPR,gkioxari2019mesh,kundu20183d,zakharov2020autolabeling,zakharov2021singleshot}, leading to remarkable improvements in terms of monocular 3D reconstruction fidelity.

Many existing works~\cite{cmrKanazawa18,saito2019pifu,gkioxari2019mesh,kundu20183d,zakharov2020autolabeling,zakharov2021singleshot} are limited in their applicability -- in particular due to their imposed data and supervision requirements: 
The majority of works require multiple views and non-occluded visibility of the same physical object, near-perfect camera pose information, and the object of interest being central, at high resolution, and thus the most prominent content of the image. Due to a lack of real-world datasets providing such features (with the notable, recently released exception of \cite{Reizenstein_2021_ICCV}), % satisfying these required high-quality annotations, 
an overwhelming number of methods have only shown experimental results on synthetic datasets, and thus under perfect data conditions, or require large datasets of CAD models to construct a shape prior. When applied to real data the existing domain gap becomes evident, typically leading to major performance degradation. %Noteworthy exceptions are \cite{}, which leverage STILL TO FILL.

Our work investigates the limits of novel view synthesis from monocular, single-image real-world data. We focus on street-level imagery where objects like cars have high variability in scale and can be very small compared to the full image resolution. Also, such objects are often occluded or may suffer from motion blur as a consequence of the data acquisition setup. We only consider a single-view object scenario during both training and inference, \ie, we do not impose constraints based on multiple views of the same object. For supervision, we limit our method to learning only from machine-generated predictions, leveraging state-of-the-art and off-the-shelf, image-based 3D object detection~\cite{Simonelli_2019_ICCV,Kumar_2021_CVPR} and instance/panoptic segmentation algorithms~\cite{He_2017_ICCV,Kirillov_2019_CVPR,Porzi_2021_CVPR}, respectively. This data setting also enables us to benchmark our results on existing autonomous driving research datasets~\cite{geiger2013vision,caesar2020nuscenes,Neuhold_2017_ICCV}. However, the absence of human quality control requires our method to cope with label noise introduced by machine-predictions from monocular 3D object detectors (itself addressing an ill-posed problem), and imperfect instance segmentation masks. 

Our proposed method follows an encoder/decoder architecture trained on images with machine-predicted 3D bounding boxes and corresponding 2D panoptic segmentation masks per image. The encoder learns to transform a training sample from its actual (arbitrary) pose and scale representation into two canonical, object-centric encodings representing shape and appearance, respectively. 
The decoder translates the object's shape and appearance codes into an object-centric, implicit radiance field representation, which provides occupancy and color information for given 3D points and viewing directions in object space.
Our training procedure benefits from the segmentation mask to gather information about the object's foreground pixels and to cope with potential occlusions, while it leverages the pose information provided by the 3D bounding box to enforce the object-centric representation.
At test time, we further optimize predicted latent codes to fit the representation tightly to the given input image by using a photometric loss formulation.
Ultimately, our architecture can learn strong implicit priors that also generalize across different datasets. We provide insightful experimental evaluations and ablation studies on challenging real-world and controllable synthetic datasets, defining a first state of the art for the challenging training setting that we consider.
In summary, our key contributions and differences with respect to existing works are:
\begin{itemize}
    \item We introduce novel view synthesis based on 3D object priors, learnt from only single-view, in-the-wild observations where objects are potentially occluded, have large variability in scale, and may suffer from degraded image quality. %In contrast to the majority of related works, w
    We neither leverage multiple views of the same object, nor utilise large CAD model libraries, or build upon specific, pre-defined shape priors.
    \item We successfully exploit machine-generated, 3D bounding boxes and panoptic segmentation masks and thus imperfect annotations for learning an implicit object representation that can be applied to novel view synthesis on real-world data. Most previous works have shown experiments on synthetic data or require the object of interest to be non-occluded and the main content of the image (except for \cite{Henzler_2021_CVPR} leveraging masks from \cite{He_2017_ICCV}).
    %\todo{Check \cite{Henzler_2021_CVPR} and \cite{Reizenstein_2021_ICCV}, maybe also machine generated}
    \item Our method efficiently encodes shape- and appearance properties for the objects of interest, which we are able to decode to a novel view in a single shot, and optionally fine-tune further at test time. This enables corrections from potential domain shifts and to generalise across different datasets, which has not been demonstrated so far.
\end{itemize}

\section{Related works}
\label{sec:related}

\myparagraph{3D reconstruction from a single image}
The task of extracting 3D information from a single image, also known as ``inverse graphics'' has received considerable attention in recent years.
Several works focus on reconstructing the shape, or the shape and appearance of a single object per image~\cite{Henderson_2020_CVPR,cmrKanazawa18,saito2019pifu,Wu_2020_CVPR}, while others attempt to extract multiple objects per image~\cite{EngelmannWACV17_samp,gkioxari2019mesh,kundu20183d,zakharov2020autolabeling} or to build a holistic representation of an entire scene~\cite{zakharov2021singleshot,dahnert2021panoptic}.
All of these approaches use differentiable rendering to formulate a reconstruction cost to compare the predicted 3D model to the 2D image while differing in the specific representation used to encode the 3D model.
Common choices here include 3D meshes~\cite{gkioxari2019mesh,Henderson_2020_CVPR,cmrKanazawa18}, signed distance functions~\cite{EngelmannWACV17_samp,mueller2021completetracking,kundu20183d}, depth~\cite{Wu_2020_CVPR}, and implicit models~\cite{zakharov2021singleshot,saito2019pifu,zakharov2020autolabeling}.
The last option, \ie, implicit models, will be the focus of the next section, and is the one we adopt in our work.

Most of these approaches exploit some form of shape prior, either learned in the form of an implicit model~\cite{zakharov2021singleshot,zakharov2020autolabeling}, or constructed from some collection of template shapes~\cite{EngelmannWACV17_samp,kundu20183d}.
Either way, they utilize large libraries of CAD models %(\eg, ShapeNet~\cite{chang2015shapenet}) 
to bootstrap their networks or as the sole form of training data, incurring in a considerable domain gap when moving to real images.
Similarly, multiple views of the same object are generally required at training time, further justifying the use of synthetic data.
In contrast, our method can be trained with a \emph{single view} per object, and entirely on \emph{real images}.
Among those mentioned above, the only works that overcome these limitations are those of Henderson~\etal~\cite{Henderson_2020_CVPR} and Wu~\etal~\cite{Wu_2020_CVPR}.
These, however, utilize high-resolution, unobstructed, and generally clean views of the objects of interest (\eg, well-lit frontal shots of faces~\cite{Wu_2020_CVPR}).
In contrast, our method is trained and validated on occluded and/or low-resolution images.

\myparagraph{Differentiable rendering and implicit models} A common denominator of many inverse graphics formulations is the use of ``differentiable rendering''.
Differentiable rendering encompasses a large array of techniques to produce 2D views of some 3D model, while also allowing for the back-propagation of gradients from the image domain to the model's parameters.
The earliest such formulations included forms of ``differentiable rasterization''~\cite{chen2019learning,liu2019soft,loper2014opendr}; \ie, back-propagable extensions of traditional rendering algorithms for (textured) 3D meshes.
More recently, approaches based on volumetric rendering have found great success, in particular when coupled with so-called ``implicit models''~\cite{chibane2020implicit,lombardi2019neural,mildenhall2020nerf,Niemeyer2020CVPR,sitzmann2019scene,yu2021pixelnerf}.
\newline\indent Implicit models represent objects or scenes as functions (\ie, neural networks), which map from 3D points to some local set of properties of the entity being modeled, \eg, whether the entity occupies the given point~\cite{mescheder2019occupancy}.
While some approaches focus solely on encoding shape~\cite{mescheder2019occupancy,park2019deepsdf,genova2020local,jiang2020lig}, others~\cite{mildenhall2020nerf,yu2021pixelnerf,lombardi2019neural,Niemeyer2020CVPR,sitzmann2019scene} have shown success in encoding shape and appearance together, meaning that i) they can produce photo-realistic renderings of the object they encode; and ii) can be trained using only 2D images.
In our work we follow the Neural Radiance Fields (NeRF) formulation of Mildenhall~\etal~\cite{mildenhall2020nerf}, which belongs to the latter category.
Differently from NeRF, our model can generalize across multiple objects, and can synthesize novel views of an object given a single input image. The capability of single-view, novel view synthesis has investigated before with Scene Representation Networks~\cite{sitzmann2019scene}, ShaRF~\cite{rematasICML21} and pixelNeRF of Yu~\etal~\cite{yu2021pixelnerf}. In \cite{sitzmann2019scene}, a scene presentation is learnt that can be fine-tuned to a test scene with given camera poses. In ShaRF, a shape generation network is pre-trained on synthetic data and optimized via Generative Latent Optimization~\cite{bojanowski2018optimizing}, whereas in this work, we do not use any geometric information. As pixelNeRF leverages local image features to synthesis novel views, it is (in contrast to ours) trained on at least two views of the same instance. In the recent work FiG-NeRF~\cite{Xie_2021_3DV}, Xie~\etal introduce a 2-component, deformable neural radiance field for jointly modeling object categories and a foreground/background segmentation. In \cite{Ost_2021_CVPR}, Ost~\etal learn a scene graph to represent automotive data enabling novel views. However, their method requires video data and remains in the scenario of over-fitting the model to single scenes. The latter limitation is addressed in \cite{Reizenstein_2021_ICCV} and \cite{Henzler_2021_CVPR}, where large-scale posed video data is leveraged to learn object-category specific generalization for novel view rendering. In contrast to all these methods, we use solely a single view of each object instance, allowing us to leverage large-scale, unstructured data at training time. Related to ours, CodeNeRF~\cite{Jang_2021_ICCV} leverages a neural radiance field that learns to disentangle latent embeddings for shape and appearance in an object-specific way. However, since CodeNeRF works as an auto-decoder architecture, it requires optimizing shape and appearance codes at test time to photometrically match a given input image before novel view synthesis can be run. \Ours does not have this limitation, because it can regress shape and appearance codes directly via the encoder, yet offering the possibility of refining the object encoding at test-time.
We finally refer to~\cite{tewari2021advances} for a survey about recent advances in neural rendering.

\section{Method}
Given a \emph{single} input image, our goal is to encode each 3D object that is present in the scene into a compact representation that allows to, \eg, efficiently store the objects into a database and re-synthesize them from different views/contexts in a later stage. While this problem has been already addressed in the past~\cite{Jang_2021_ICCV,yu2021pixelnerf}, we focus on a more challenging scenario when it comes to training such an encoder. As opposed to the vast majority of methods that assume to have access to at least a second view of the same object instance when training such an encoder, we focus on addressing the more challenging setting where object instances can be observed from a single view only. Moreover, no other prior knowledge about the object's geometry (\eg, CAD models, symmetries, etc.) is exploited. Finally, we train our model using real-world images that have not been curated for the specific task at hand. \Eg, we might leverage 3D object detection datasets to train, where images contain multiple, possibly occluded object instances, each having possibly different scales and thus resolutions.

To be able to train the encoder in the underconstrained scenario mentioned above, we take advantage of pre-trained instance or panoptic segmentation algorithms to identify in the image 2D pixels belonging to the same object instance as well as pre-trained monocular 3D object detectors in order to get a prior about the objects' poses in 3D space.
Accordingly, both at training and test time, we assume to get for each image a set of 3D bounding boxes with associated 2D masks, which represent the detected object instances, and information about camera calibration.

By leveraging the information about the object 3D bounding box, we can disentangle the object representation from the actual object pose and scale. Indeed, we obtain a normalized, object-centric encoding, which is factored into a shape and an appearance component. Akin to a conditional NeRF model, the shape code is used to condition an occupancy network, which outputs a density given a 3D point in normalized object space and the appearance code is used to condition an appearance network that provides a RGB color given a 3D point and a viewing direction in normalized object space. The two networks yield an implicit representation of the 3D object.

\subsection{Preliminaries}\label{ssec:preliminaries}

\begin{figure}
    \centering
    \includegraphics[width=\columnwidth]{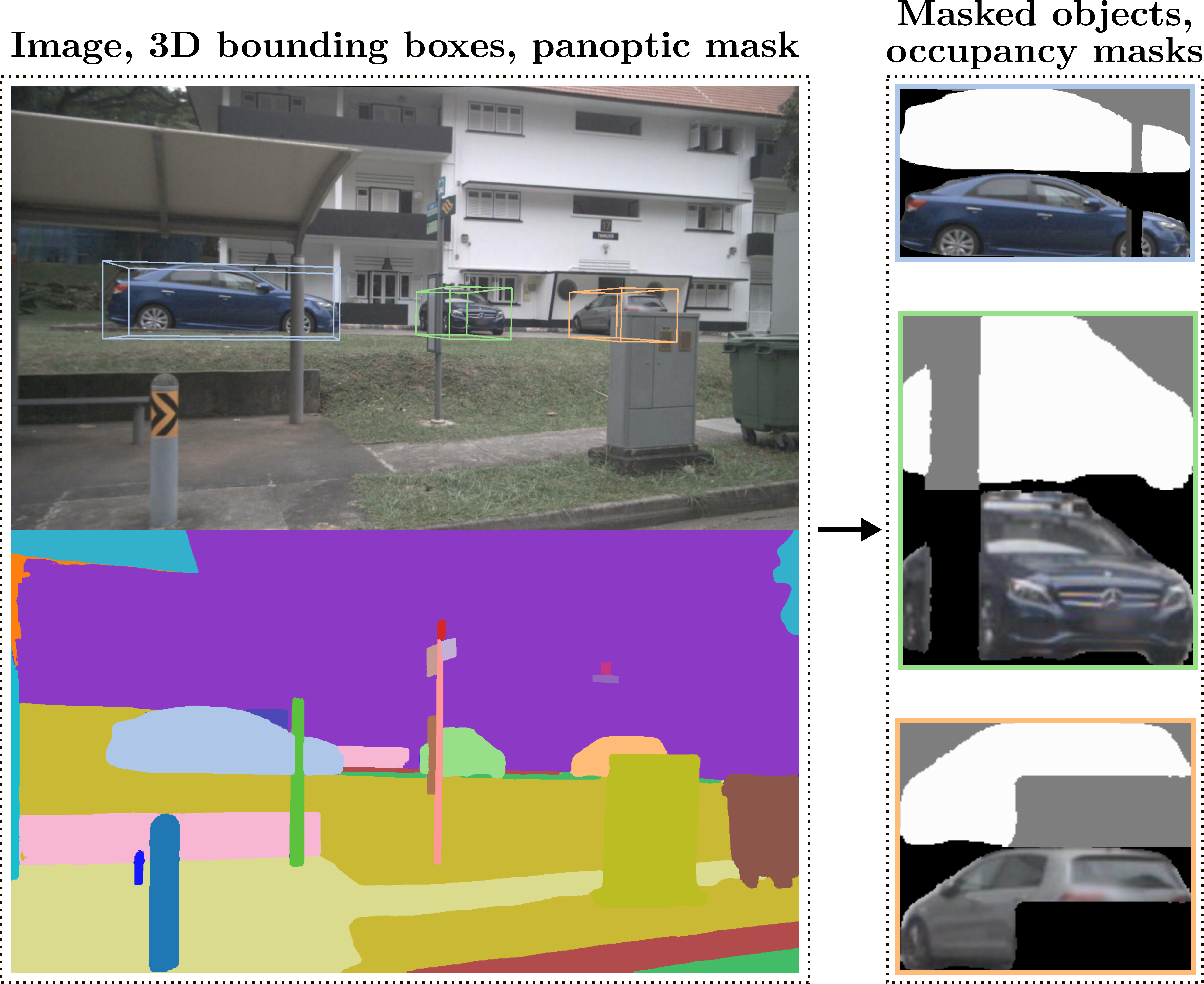}
    \caption{Pre-processing step: first, we use pre-trained models to detect the objects of interest in 3D and segment the image. Then, we crop per-object views and compute their occupancy masks (white: foreground, black: background, grey: unknown).}
    \label{fig:data_generation}
\end{figure}

\myparagraph{Image $I$}
Given a 2D image where multiple objects of interest are present, we run a 3D object detector along with panoptic segmentation in order to extract for each object instance a 3D bounding box and an instance mask (see Fig.~\ref{fig:data_generation}). The bounding box and the mask are used to produce a masked 2D image $I$ of the detected object instance, fitting a fixed input resolution. In addition, the 3D bounding box captures extent, position, and rotation of an object in camera space, while the segmentation mask provides per-pixel information about possible occlusions \wrt other objects in the scene.
The RGB color of pixel $u\in\mathcal U$ in image $I$ is denoted by $I_u\in\mathbb R^3$, where $\mathcal U$ represents its set of pixels. 

\myparagraph{Normalized Object Coordinate Space $\mathcal O$}
Each object instance has an associated 3D bounding box $\beta$ that identifies a rectangular cuboid in camera space describing the pose and extent of the associated object.
%A 3D bounding box $b\in\mathcal B$ identifies a rectangular cuboid in camera space describing the pose and extent of the associated object. 
The 3D points contained in a 3D bounding box $\beta$ can be mapped via a diffeomorphism to the (centered) unit cube $\mathcal O\coloneqq\left[-\frac{1}{2},\frac{1}{2}\right]^{3}$ called Normalized Object Coordinate Space (NOCS). Indeed, every 3D bounding box can be translated, rotated and scaled into a unit cube. In light of this fact, we will use directly $\beta$ to represent the aforementioned diffeomorphism and, hence, to map points from camera space to the NOCS. %\todo{PK: It would be great to have a visualisation for the NOCS}

\myparagraph{Object-Centric Camera $\gamma$}
Each image $I$ depicting a 3D scene has an associated camera denoted by $\rho$. Camera $\rho$ maps
%$\rho:\mathcal U\times\mathbb R_+\to\mathbb R^3$ 
 pixels $u\in\mathcal U$ to unit-speed rays in camera space denoted by $\rho_u:\mathbb R_+\to\mathbb R^3$, where $\rho_u(t)$ gives the 3D point along the ray at time $t$. 
By leveraging the bounding box $\beta$ of a given object, we can map each ray $\rho_u$ from camera space to NOCS yielding an object-centric ray $\gamma_u$. Specifically, $\gamma_u$ is a unit speed reparametrization of the remapped ray $\beta\circ\rho_u$. We refer $\gamma$ as the object-centric camera for a given object.

\myparagraph{Occupancy Mask $Y$}
We use panoptic segmentation to produce a 2D occupancy mask $Y$ associated with an object's image $I$.
An occupancy mask $Y$ provides for each pixel $u\in\mathcal U$ a class label $Y_u\in\{+1,0,1\}$. Foreground pixels, \ie, pixels belonging to the object instance mask, are assigned label $+1$. Background pixels, \ie, pixels that are not occluding the object of interest, are assigned label $-1$. Pixels for which it is not possible to determine whether they occlude the object or not are assigned label $0$. A pixel is assigned the background label if it belongs to a semantic category that is not supposed to occlude the object of interest (\eg, for car, we have sky, road, sidewalk, \etc). See Fig.~\ref{fig:data_generation} for an example.

\subsection{Architecture Overview}

\begin{figure*}
\begin{center}
\includegraphics[width=0.95\textwidth]{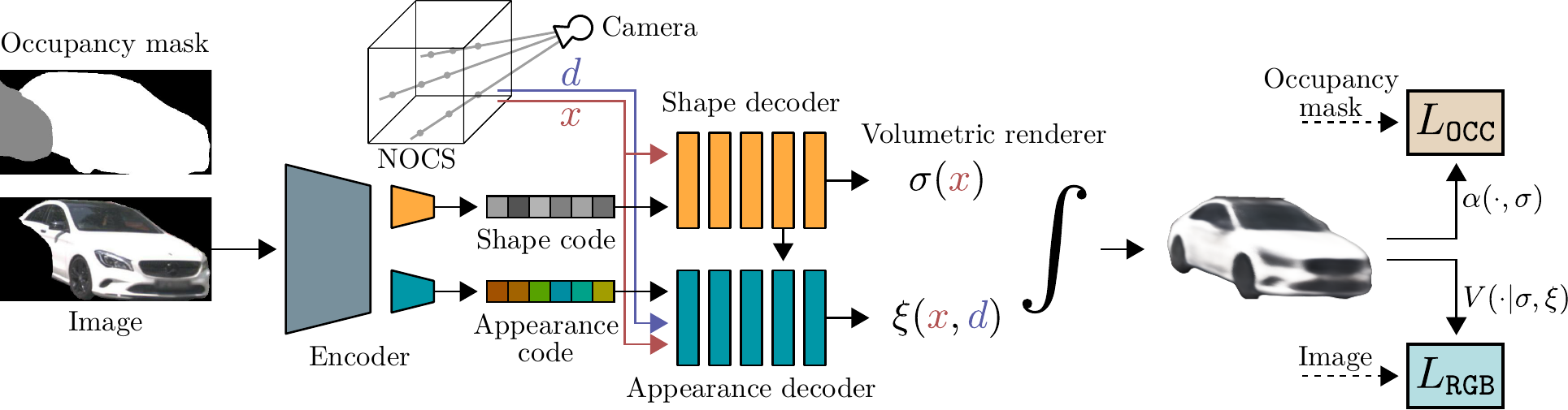}
\vspace{-1em}
\end{center}
\caption{
Given an RGB image with a corresponding 3D object bounding box and occupancy mask, our autoencoder learns to encode shape and appearance in separate codes. These codes condition individual decoders to re-render the input image for the given view.  
}
\vspace{-10pt}
\label{fig:overview}
\end{figure*}

We provide an overview of our architecture in Fig.~\ref{fig:overview} and provide a description below.

\newcommand{\shape}{{\phi_\mathtt S}\xspace}
\newcommand{\appearance}{{\phi_\mathtt A}\xspace}

\myparagraph{Inputs $(I,\gamma,Y)$}
Our architecture takes as input the image $I$ of an object that has been detected, the corresponding camera $\gamma$ in NOCS, which has been derived by exploiting the information about the object's 3D bounding box, and the occupancy mask $Y$ obtained by leveraging panoptic segmentation. Details about $I$, $\gamma$ and $Y$ have been provided in Sec.~\ref{ssec:preliminaries}. Examples of object images, occupancy masks and 3D bounding boxes are given in Fig.~\ref{fig:data_generation}.

\myparagraph{Shape and Appearance Encoder $\Phi_\mathtt{E}$} 
We encode an input image $I$ depicting a given object of interest into a \emph{shape} code $\shape$ and an \emph{appearance} code $\appearance$ via a neural network $\Phi_\mathtt{E}$; \ie, $(\shape,\appearance)\coloneqq\Phi_\mathtt{E}(I)$. 
The encoder comprises a CNN feature extractor that outputs intermediate features that are fed to two parallel heads, responsible for generating the shape and appearance code, respectively.
Implementation details of the encoder and decoders that follow can be found in the supplementary material.

\myparagraph{Shape Decoder $\Psi_\mathtt S$}
The shape code $\shape$ is fed to a decoder network $\Psi_\mathtt{S}$, which \emph{implicitly} outputs an occupancy network $\sigma$; \ie, $\sigma\coloneqq\Psi_\mathtt S(\phi_\mathtt S)$.
The occupancy network $\sigma:\mathcal O\to\mathbb R_+$ outputs a density for a given 3D point $x\in\mathcal O$ expressed in NOCS. 

\myparagraph{Appearance Decoder $\Psi_\mathtt A$}
As opposed to the shape decoder, the appearance decoder $\Psi_\mathtt A$ takes in input both shape and appearance codes and \emph{implicitly} outputs an appearance network $\xi$, \ie $\xi\coloneqq\Psi_\mathtt A(\phi_\mathtt A, \phi_\mathtt S)$. The appearance network $\xi:\mathcal O\times\mathbb S^2\to\mathbb R^3$ outputs an RGB color for a given 3D point $x\in\mathcal O$ and a viewing direction $d$ on the unit 3D sphere $\mathbb S^2$.

\myparagraph{Volume Renderer $V$}
The occupancy network $\sigma$ and the appearance network $\xi$ form a \emph{radiance field} representing an object in NOCS.
We can compute the color associated with $u$ by rendering the object-centric ray $\gamma_u$ using the approach proposed in~\cite{mildenhall2020nerf}. However, since we are interested in modelling only the object of interest, the object-centric ray is limited to points intersecting $\mathcal O$. This yields the following volume rendering formula:\footnote{The formula differs at first sight from the one in~\cite{mildenhall2020nerf}, but the two become equivalent after computing the time derivative $\dot\alpha_t$.}
\[
V(\gamma_u|\sigma,\xi)\coloneqq-\int_{a_u}^{b_u} \dot\alpha_t(\gamma_u, \sigma)\xi(\gamma_u(t), d_u ) dt\,,
\]
where $[a_u,b_u]$ is the time-window where $\gamma_u$ intersects $\mathcal O$ and $d_u\in\mathbb S^2$ is the unit velocity of $\gamma_u$.
Moreover, $\dot\alpha_t$ denotes the time derivative of the accumulated transmittance along the ray $\gamma_u$ in the range $[a_u,t]$ defined as
\[
\alpha_t(\gamma_u,\sigma)\coloneqq \exp\left(-\int_{a_u}^t\sigma(\gamma_u(s)) ds\right)\,.
\]
The integral in the volume renderer $V$ can be solved with a quadrature rule by leveraging sampled points along the ray (see, \cite{mildenhall2020nerf} for more details).

\subsection{Training}
To train our architecture we rely on two loss terms: a photometric loss and an occupancy loss.
We provide the losses for a given training example $\Omega=(I,\gamma,Y)$ comprising the image $I$, the occupancy mask $Y$, the object-centric camera $\gamma$.
Moreover, we assume that the radiance field $(\sigma,\xi)$ for the object has been computed from $I$ using the encoder $\Phi_\mathtt E$ and the decoders $\Psi_\mathtt S$ and $\Psi_\mathtt A$.
Finally, we denote by $\Theta$ all learnable parameters involved in the architecture.

\myparagraph{Photometric Loss $L_\mathtt{RGB}$} The photometric loss term resembles an autoencoder loss, for it forces the model to fit the input it is given after encoding it into a shape and appearance code with $\Phi_\mathtt E$, decoding it into an object radiance field by using $\Psi_\mathtt S$ and $\Psi_\mathtt A$, and finally rendering it with the volume renderer $V$. The loss is formally defined as follows
\[
L_\mathtt{RGB}(\Theta|\Omega)\coloneqq\frac{1}{|\mathcal W_\mathtt {RGB}|}\sum_{u\in\mathcal W_\mathtt{RGB}}\Vert I_u - V(\gamma_u|\sigma, \xi) \Vert^2\,,
\]
where $\mathcal W\subset\mathcal U$ contains only foreground pixels; \ie, $Y_u=+1$ whose object-centric rays $\gamma_u$ intersect $\mathcal O$.

\myparagraph{Occupancy Loss $L_\mathtt{OCC}$} We use panoptic segmentation to infer whether a pixel is a foreground, background or unknown pixel. This information is encoded in the occupancy mask $Y$, which is used to directly supervise the accumulated transmittance component $\alpha$ of the volume rendering equation $V$.
 Indeed, $\alpha(\gamma_u,\sigma)\coloneqq\alpha_{b_u}(\gamma_u,\sigma)$ represents the probability that the object does not intersect ray $\gamma_u$, or in other terms that $u$ is potentially a background pixel. Similarly $1-\alpha(\gamma_u,\sigma)$ is the probability of $u$ to be a foreground pixel. We can therefore implement a classification loss directly on the accumulated transmittance as follows:
\[
L_\mathtt{OCC}(\Theta|\Omega)\coloneqq-\frac{\sum_{u\in\mathcal W_{\mathtt {OCC}}}\log\left[Y_u(\frac{1}{2}-\alpha(\gamma_u,\sigma))+\frac{1}{2}\right]}{|\mathcal W_\mathtt{OCC}|}\,.
\]
where $\mathcal W_\mathtt{OCC}\subset\mathcal U$ contains only foreground or background pixels; \ie, $Y_u\neq 0$, with rays $\gamma_u$ intersecting $\mathcal O$.

\myparagraph{Final Loss $L$} The final loss that we use to train our network is a linear combination of the two losses described above:
\[
L(\Theta|\Omega)=L_\mathtt{RGB}(\Theta|\Omega)+\lambda L_\mathtt{OCC}(\Theta|\Omega)\,,
\]
where the occupancy loss is  modulated with a hyperparameter $\lambda\geq 0$.

\begin{figure*}[ht!]
\begin{center}
\includegraphics[width=2.1\columnwidth, ]{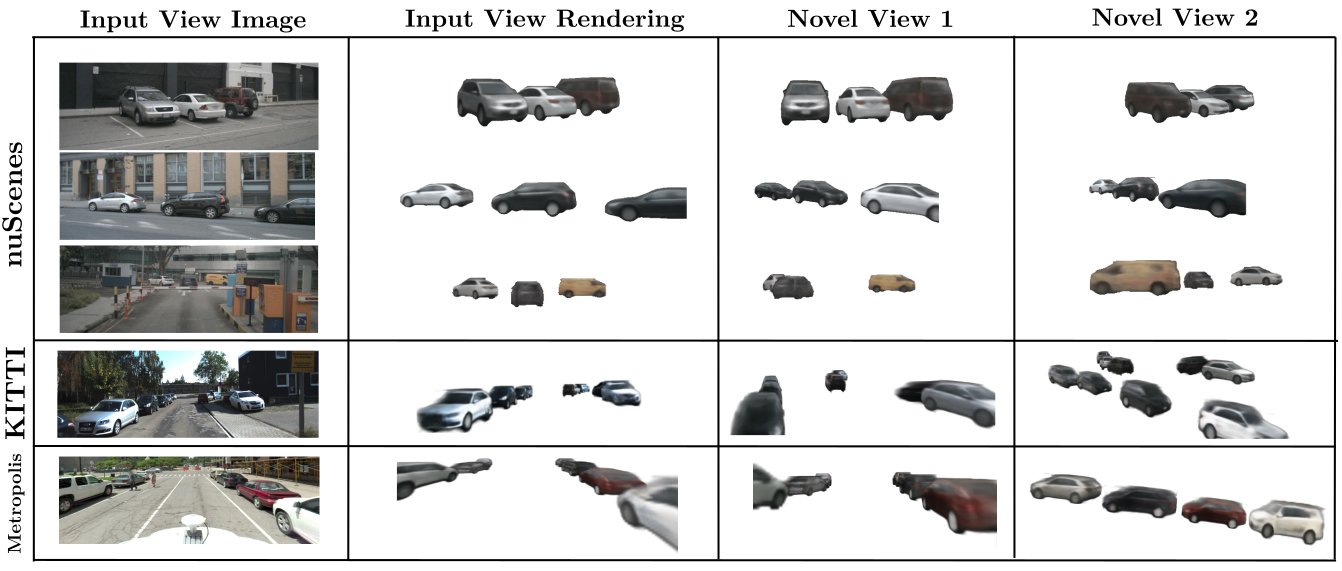}
\end{center}
\vspace{-15pt}
\caption{
Full scene novel view synthesis from single unseen images for nuScenes (top), KITTI (middle), and Mapillary Metropolis (bottom) after training our model exclusively on nuScenes test data. Please note the high-fidelity reconstruction results obtained for objects of different scale, aspect ratio, and image quality.
}
\vspace{-10pt}
\label{fig:view_synthesis_qualitative}
\end{figure*}

\subsection{Test-Time Optimization}
Our method enables a forward encoding of an object at test-time due to the presence of an ad-hoc encoder $\Phi_\mathtt E$. Nevertheless, we can further refine the regressed codes or even the object's prior pose to make the output more robust, for instance, to domain shifts in the object's appearance or errors in the predicted 3D bounding box. To this end, we keep optimizing our loss $L$ at test time, but with the object's shape/appearance codes $(\phi_\mathtt S, \phi_\mathtt A)$ and the 3D bounding box $\beta$ being regarded as variables to be optimized. Since the optimization requires a good initial estimate to converge towards a good solution, we use the initial 3D bounding box prediction from the 3D object detector and the object's encoding provided by our encoder $\Phi_\mathtt E$ to initialize the variables. Our formulation allows also to optimize a subset of those variables by keeping the other fixed (\eg, fine-tune appearance only by optimizing $\phi_\mathtt A$, while keeping $\phi_\mathtt S$ and $\beta$ fixed). It is worth mentioning that in a monocular setting optimizing the bounding box $\beta$ is not well-defined, because of the scale-depth ambiguity. In practice, we keep the size component of the bounding box fixed to the one regressed by the 3D object detection and optimize only the bounding box pose; \ie, rotation and translation.

\section{Experiments}\label{sec:exps}

We quantitatively evaluate our approach for the task of novel view synthesis from a single view on the nuScenes dataset \cite{caesar2020nuscenes}, and on the SRN-Cars synthetic car dataset introduced in \cite{sitzmann2019scene}. Finally, we also evaluate our model trained on nuScenes data on images taken from the KITTI~\cite{Geiger2012CVPR} and Mapillary Metropolis\footnote{\url{https://www.mapillary.com/dataset/metropolis}} datasets, respectively. In Fig.~\ref{fig:view_synthesis_qualitative} we provide reconstruction results together with synthesised, novel views after training on nuScenes for 1) nuScenes validation (top), 2) KITTI validation (middle), and 3) Mapillary Metropolis validation (bottom). Please note that the model has never seen any data from KITTI or Metropolis during training.  %of ShapeNet \cite{chang2015shapenet} renderings for cars.

\myparagraph{Baselines}
We compare quantitatively and qualitatively to PixelNeRF \cite{yu2021pixelnerf} on the task of one-view, 2D-supervised reconstruction. 
For experiments on nuScenes, we extend their method to support training only on foreground and background pixels, and transform the camera system into the normalized object space in order to leverage 3D object annotations. 
As pixelNeRF is trained in a multi-view setup, we provide an additional view during training time leveraging provided tracking annotations. In contrast, we train our model using only a single observation of the same instance. 

\myparagraph{Metrics} We report the standard image quality metrics PSNR (Peak Signal to Noise Ratio) and SSIM (Structural Similarity Index Measure) \cite{Wang04imagequality} for all evaluations. Furthermore, we include LPIPS \cite{zhang2018perceptual} and FID \cite{FID} scores to more accurately reflect human perception.

\myparagraph{Implementation Details}
Similar to PixelNeRF~\cite{yu2021pixelnerf}, our image encoding uses a ResNet34 backbone pre-trained on ImageNet, while each decoder consists of five fully-connected residual blocks.
For an in-depth description of our architecture, we refer to the supplementary material.

\begin{comment}
For our shape and appearance encoder $\Phi_\mathtt{E}$, an input image is first passed through the first 3 layers of ResNet34 to obtain shared intermediate features. Two separate branches each consisting of the fourth and fifth level of ResNet34 followed by an adaptive average pooling then output the final shape and appearance codes of size 128 each.
The shape and appearance decoder both consist of 5 ResNet blocks with hidden dimensions of 128, where each the residual of each block is conditioned on additional input: We feed the first 6 frequencies of a positional encoding of the query coordinates in NOCS via independent linear layers into each block for both decoders. Additionally, we condition the appearance decoder on the intermediate shape features of the fourth block of the shape decoder as well as on the camera viewing direction $d$. This informs the appearance part about the object shape and allows for view-dependent effects.
\end{comment}

\subsection{Evaluation on nuScenes}
\label{ssec:nuScenesEval}

The nuScenes dataset is a large-scale driving dataset with 3D detection and tracking annotations for 7 object classes. It contains 700 training, 150 validation, and 150 test sequences, comprising 168k training images, 36k validation images, and 36k test images. As this dataset is commonly used for perception tasks in autonomous driving research, we pre-process the data to make it suitable for the task of novel view synthesis: 
We filter for sequences at daytime (provided as meta-information) and we run a pre-trained 2D panoptic segmentation model \cite{Porzi_2021_CVPR} as nuScenes does not provide 2D segmentation masks. 

We match provided 3D bounding box annotations with the resulting instance masks and categorize the panoptic results into foreground (visible part of the object), background (non-occluding semantic categories like street, sky, sidewalk), and unknown regions (potentially occluding categories like people, vehicles or vegetation) as we do not rely on depth information in order to resolve occlusions. Furthermore, we filter for sufficiently visible instances and use tracking information for evaluation purposes. For training, we select from each instance a single view at random to train on (two views for the baseline) and evaluate on a pre-determined subset of 10k pairs of views of car instances in the validation split. We refer to the supplementary material for details about the data generation process.

\begin{figure}
\begin{center}
\includegraphics[width=.99\columnwidth]{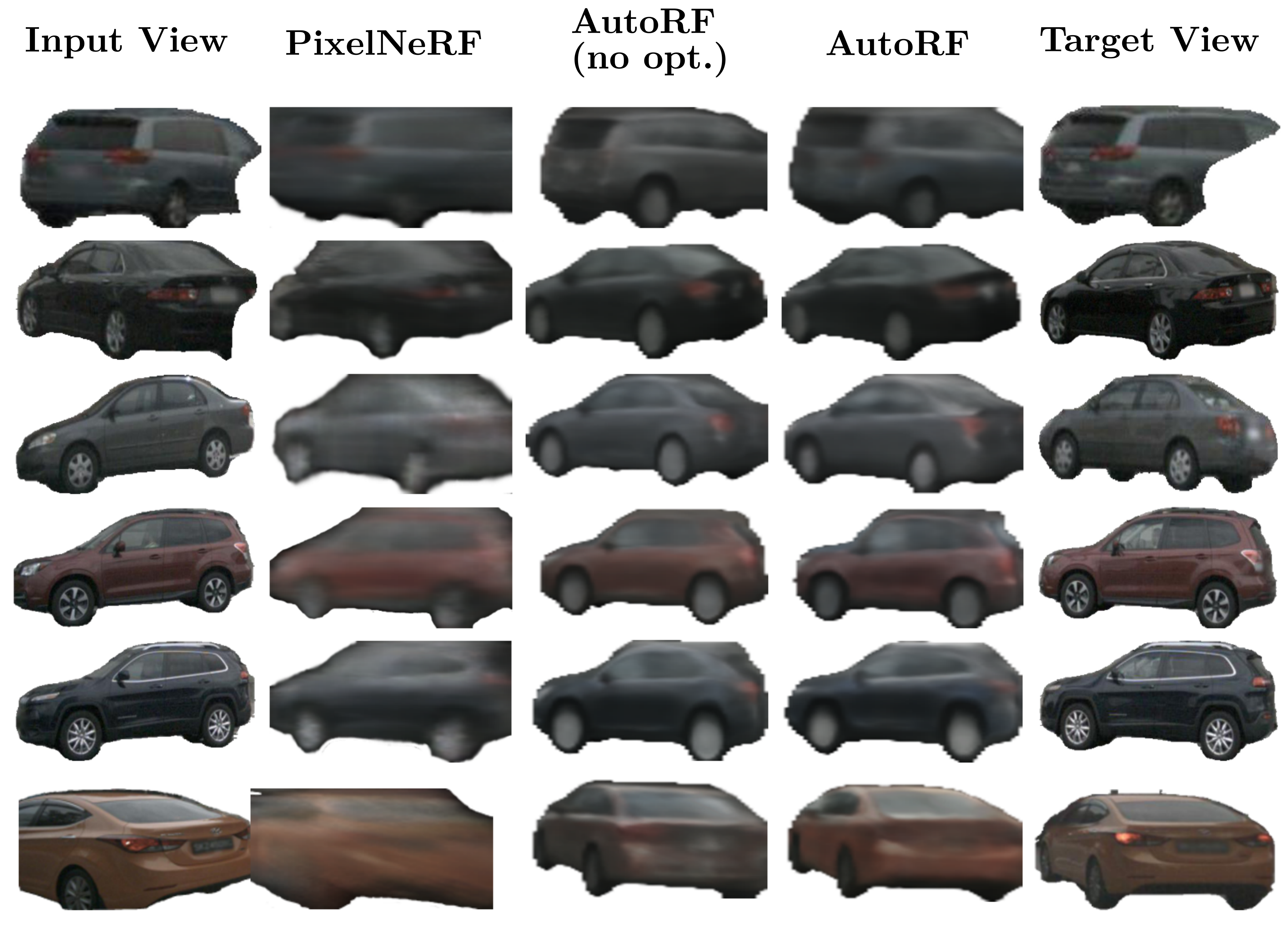}
\end{center}
\vspace{-10pt}
\caption{
Qualitative comparison on NuScenes: novel view synthesis of single instances.
}
\label{fig:nusc_results1}
%\vspace{-5pt}
\end{figure}

\begin{figure}
\centering
\resizebox{\columnwidth}{!}{
\begin{tabular}{cc}
    \centering
    \begin{minipage}{1.20\columnwidth}
        \begin{tikzpicture}
        \begin{axis}[
            xlabel={Novel View Rotation Delta [\textdegree]},
            ylabel={LPIPS~$\downarrow$},
            xmin=-10, xmax=190,
            ymin=0.175, ymax=0.300,
            xtick={0,20,40,60,80,100,120,140,160,180},
            ytick={0.175, 0.200, 0.225, 0.250, 0.275, 0.300},
            legend pos=north west,
            ymajorgrids=true,
            grid style=dashed,
            label style={font=\Large},
        ]
        
        \addplot[
            color=color2!60!black,
            mark=square,
            line width=1.5pt,
            mark size=3pt,
            ]
            coordinates {
            (0,0.1928)(20,0.2157)(40,0.2376)(60,0.2549)(80,0.2621)(100,0.2721)(120,0.2791)(140,0.2824)(160,0.2717)(180,0.2581)
            };
        
        \addplot[
            color=color1!60!black,
            mark=o,
            line width=1.5pt,
            mark size=3pt,
            ]
            coordinates {
            (0,0.2017)(20,0.2109)(40,0.2272)(60,0.2362)(80,0.2437)(100,0.2502)(120,0.2543)(140,0.2563)(160,0.2499)(180,0.2449)
            };
        
        \addplot[
            color=color3!60!black,
            mark=triangle,
            line width=1.5pt,
            mark size=3pt,
            ]
            coordinates {
            (0,0.1790)(20,0.1958)(40,0.2103)(60,0.2195)(80,0.2308)(100,0.2342)(120,0.2397)(140,0.2399)(160,0.2403)(180,0.2344)
            };
        
        \legend{PixelNeRF, \Ours (no opt.), \Ours}    
        \end{axis}
        \end{tikzpicture}
    \end{minipage} & %\quad &
    \begin{minipage}{1.20\columnwidth}
        \begin{tikzpicture}
        \begin{axis}[
            xlabel={Novel View Rotation Delta [\textdegree]},
            ylabel={PSNR~$\uparrow$},
            xmin=-10, xmax=190,
            ymin=16, ymax=21,
            xtick={0,20,40,60,80,100,120,140,160,180},
            ytick={16, 17, 18, 19, 20, 21},
            legend pos=north east,
            ymajorgrids=true,
            grid style=dashed,
            label style={font=\Large},
        ]
        
        \addplot[
            color=color2!60!black,
            mark=square,
            line width=1.5pt,
            mark size=3pt,
            ]
            coordinates {
            (0,19.83)(20,18.54)(40,17.97)(60,17.93)(80,17.55)(100,17.2)(120,16.75)(140,17.08)(160,16.92)(180,17.57)
            };
        
        \addplot[
            color=color1!60!black,
            mark=o,
            line width=1.5pt,
            mark size=3pt,
            ]
            coordinates {
            (0,19.67)(20,18.84)(40,18.31)(60,18.41)(80,18.46)(100,18.00)(120,17.74)(140,17.93)(160,18.02)(180,19.05)
            };
        
        \addplot[
            color=color3!60!black,
            mark=triangle,
            line width=1.5pt,
            mark size=3pt,
            ]
            coordinates {
            (0,20.33)(20,19.01)(40,18.44)(60,18.57)(80,18.55)(100,18.10)(120,17.85)(140,18.00)(160,18.21)(180,19.09)
            };
        
        \legend{PixelNeRF, \Ours (no opt.), \Ours}    
        \end{axis}
        \end{tikzpicture}
    \end{minipage}
    
\end{tabular}}
\caption{
    Novel view evaluation: measures of image fidelity plotted against the rotational delta between input and target view.
}
\vspace{-10pt}
\label{fig:novel_rot}
\end{figure}
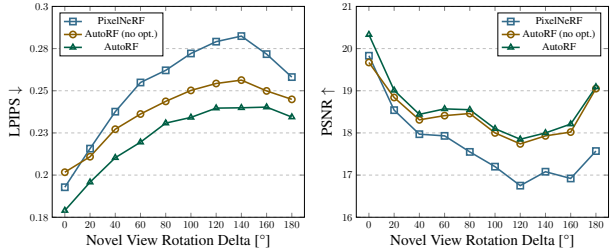

\myparagraph{Single-view synthesis results} We report our performance in comparison to state-of-the-art baselines pixelNeRF and CodeNeRF in Tab.~\ref{tab:nusc_val}.
Even without any multi-view information, our model is able to synthesize higher quality results compared to the baseline trained with multi-view information. Test-time optimization allows the model to recover instance-specific details while preserving shape and overall appearance. 
In Fig.~\ref{fig:nusc_results1} we demonstrate qualitatively that our model produces overall sharper results and more natural shapes, while pixelNeRF struggles to synthesize views that are significantly different from the input view. We quantify this observation by plotting PSNR and LPIPS values against the rotational difference between input and target view in Fig.~\ref{fig:novel_rot}: While the performance of the models evaluated on views close to the input view is similar, pixelNeRF degrade significantly with increasing change of perspective but to the maximum rotational error, where the models can leverage similarities to the input view (\eg, a car seen from the left and right side).
\begin{table}[bht]
\resizebox{\columnwidth}{!}{
\begin{tabular}{@{}lcccc@{}}
nuScenes cars       & PSNR ↑    & SSIM ↑    & LPIPS ↓   & FID ↓     \\ 
\toprule
pixelNeRF~\cite{yu2021pixelnerf}  & 18.25  & 0.459    & 0.236     & 160.60\\
CodeNeRF~\cite{Jang_2021_ICCV}  & 18.44  & 0.462    & 0.241     & 146.32\\
\midrule
\Ours{} (no opt.) on test        & 18.69 & 0.479  & 0.227    &    \textbf{138.23} \\
\Ours{} on test   & \textbf{18.94}   & \textbf{0.491}   & \textbf{0.223}   & 145.10 \\
\bottomrule
\end{tabular}}
\caption{
Overview of novel-view synthesis results on nuScenes cars from the validation set.
} % \caption
\label{tab:nusc_val}
\end{table}
\vspace{-10pt}

\subsubsection{Shape reconstruction quality}
We additionally evaluate our methods' shape reconstruction quality on the nuScenes validation split by comparing the resulting depth renderings against ground truth (GT) object LiDAR points. We crop the LiDAR recordings according to the oriented GT 3D bounding-box annotations, remove points in the lower 10\% of the bounding box (to exclude LiDAR points belonging to the street), and finally evaluate on samples with at least 20 remaining points. Tab.~\ref{tab:depth_err} shows that our model trained on single views and from auto-generated 3D detection and segmentation results produces more precise surfaces in terms of L1 and RMSE metrics compared to the pixelNeRF baseline trained with GT annotations and multiple views per instance (additional qualitative reconstructions can be found in the supplementary document).

\begin{table}[h]
\begin{minipage}{.48\columnwidth}
\centering
\resizebox{0.99\columnwidth}{!}{
\begin{tabular}{@{}lcc@{}}
nuScenes cars       & L1 ↓ & RMSE ↓  \\ 
\toprule
pixelNeRF~\cite{yu2021pixelnerf}       & 0.357 & 0.984  \\
CodeNeRF~\cite{Jang_2021_ICCV}       & 0.239 & 0.641  \\
\midrule
\Ours{} (no opt.)    & 0.209  & 0.632\\
\Ours{}   & \textbf{0.204} & \textbf{0.614}\\
\bottomrule
\end{tabular}
}
\captionof{table}{Qualitative comparison on SRN-chairs trained on single views.}
\label{tab:depth_err}
\end{minipage}
\centering
\hfill\begin{minipage}{.48\columnwidth}
\centering
\resizebox{0.99\columnwidth}{!}{
\begin{tabular}{@{}lcc@{}}
Avg. perturb.     & PSNR ↑ & LPIPS ↓    \\ 
\toprule
0$\degree$/ 0cm & \textbf{18.95} & \textbf{0.210}   \\
5$\degree$/ 10cm  & 18.67 & 0.216   \\
10$\degree$/ 20cm  & 17.95 & 0.269 \\
20$\degree$/ 40cm  & 16.83 & 0.348  \\
\bottomrule
\end{tabular}
}
\captionof{table}{Novel-view synth. of \Ours{} trained with perturbed annotations on nuScenes.}
\label{tab:data_quality}
\end{minipage}
\end{table}

\subsection{Evaluation on synthetic data} We evaluate our method against the baselines on the SRN dataset introduced in \cite{sitzmann2019scene}. The SRN-Cars dataset contains 3514 samples of car renderings (based on shapes from 3D Warehouse) with a predefined split across object instances. While each model is rendered from 50 random views per object instance, we select a single random frame for training our method and CodeNeRF (and two random frames for the pixelNeRF baseline).
For each object in the test set, we evaluate novel view synthesis from 251 views sampled on an Archimedean spiral based on a single, randomly chosen view as input for our method. We refer to the supplementary for evaluation on additional categories.

\begin{figure}
\begin{center}
\includegraphics[width=.99\columnwidth]{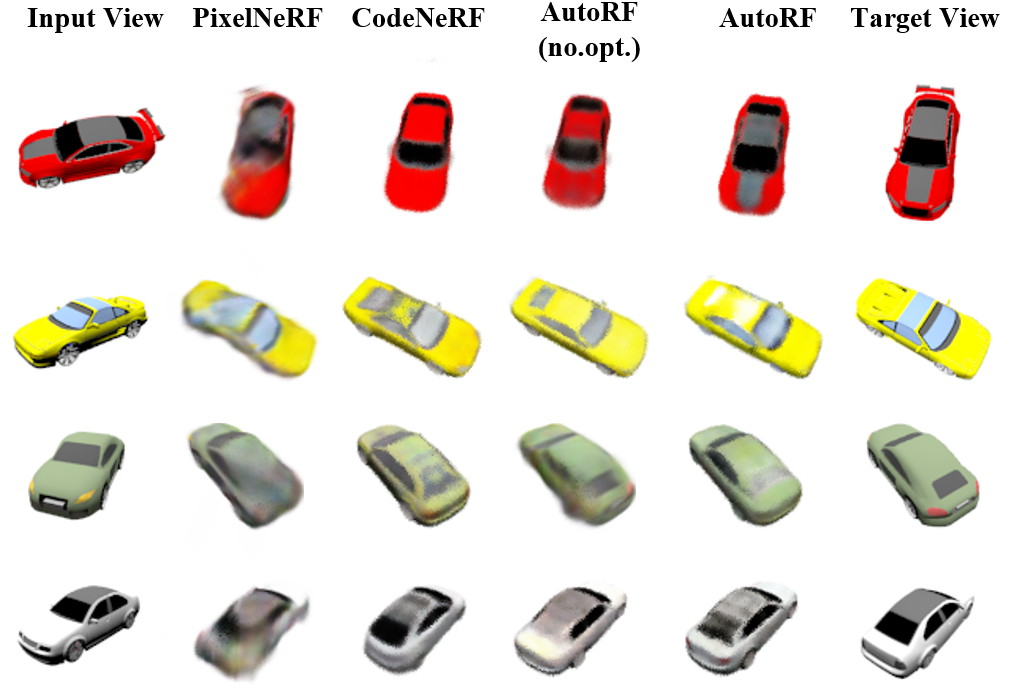}
\end{center}
\vspace{-15pt}
\caption{
Qualitative comparison on SRN-Cars dataset, illustrating our high-fidelity, single-view reconstruction results compared to the 2-view pixelNeRF baseline. 
}
\vspace{-15pt}
\label{fig:srn_cars_results1}
\end{figure}

\myparagraph{Single-view synthesis results} We compare our method in Tab.~\ref{tab:srn_cars} to pixelNeRF  trained on additional views and CodeNeRF. We notice that our model outperforms the baselines on all metrics while not requiring multi-view constraints at training time. Additionally, we provide qualitative results in Fig.~\ref{fig:srn_cars_results1} illustrating our method's high-fidelity reconstruction results, and how we are preserving fine-grained details such as differently colored roof-tops of cars.
\begin{table}[]
\centering
\resizebox{0.85\columnwidth}{!}{
\begin{tabular}{@{}lcccc@{}}
SRN-Cars       & PSNR ↑    & SSIM ↑    & LPIPS ↓   & FID ↓     \\ 
\toprule
pixelNeRF~\cite{yu2021pixelnerf}   & 19.55  & 0.847   & 0.177     & 142.9 \\
CodeNeRF~\cite{Jang_2021_ICCV}   & 18.93  & 0.844   & 0.172     & 127.1 \\
\midrule
\Ours{} (no opt.)        & 18.08 & 0.833  & 0.180    & \textbf{121.6}\\
\Ours{}   & \textbf{19.66}   & \textbf{0.860}   & \textbf{0.165}   &  122.4 \\
\bottomrule
\end{tabular}}

\caption{
Evaluation of novel-view synthesis on the SRN-Cars dataset from \cite{sitzmann2019scene}.
} % \caption
% \vspace{-10pt}
\label{tab:srn_cars}
\end{table}

\subsection{Ablations}

\myparagraph{Data quality}
In Tab.~\ref{tab:data_quality}, we report novel-view synthesis results of AutoRF trained with random perturbations of the ground truth annotations in terms of average different rotation and translation errors. We note that smaller inaccuraries have minor impact. Furthermore, we investigate the performance when we train \Ours on human-annotated data on the nuScenes train split and evaluate the results against our model fully trained on machine-annotated single-view data. 
The results are summarized in Tab.~\ref{tab:nusc_abl} and show that leveraging high-quality annotations does not significantly improve the novel view synthesis results. While PSNR and SSIM are very similar, the main improvements are gained in terms of perceptual losses (LPIPS and FID). Qualitative analyses show that the model trained on GT annotations are slightly less blurry, which we assign to the fact that inaccurate pose annotations result in imprecise ray sampling in NOC space.
\begin{table}[]
\resizebox{\columnwidth}{!}{
\begin{tabular}{@{}lcccc@{}}
nuScenes cars       & PSNR ↑    & SSIM ↑    & LPIPS ↓   & FID ↓     \\ 
\toprule
\Ours{} (no opt.) on test    & 18.69     & 0.479     & 0.227     & 138.23\\
\Ours{} (no opt.) on train   & 18.58     & 0.473     & 0.211     & \textbf{84.14 } \\
\midrule
\Ours{} on test         & 18.94     &0.491   & 0.223   & 145.10 \\
\Ours{} on train    & \textbf{18.95}    & \textbf{0.493}   & \textbf{0.210}   & 106.50  \\
\bottomrule
\end{tabular}}
\caption{
Novel-view evaluation on the nuScenes validation set.
}
% \vspace{-10pt}
\label{tab:nusc_abl}
\end{table}

\myparagraph{Domain transfer}
While trained solely on the nuScenes street-level dataset, we show qualitative results in Fig.~\ref{fig:view_synthesis_qualitative} demonstrating that the learnt object radiance field priors generalize well to novel datasets. Examples on nuScenes, KITTI, and Mapillary Metropolis show that AutoRF (no opt.) can reliably assign matching object priors and that test time optimization consistently preserves fine-grained details also in novel views.

\begin{figure}
\begin{center}
\includegraphics[width=.99\columnwidth]{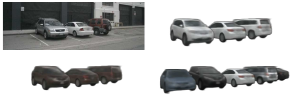}
\end{center}
\vspace{-15pt}
\caption{
Scene editing examples on nuScenes. Starting from the input view, we can change the codes of the objects and synthesize novel scene layouts.
}
\vspace{-15pt}
\label{fig:scene_editing}
\end{figure}

\myparagraph{Scene editing}
Our approach naturally decomposes an object into pose, shape, and appearance. This directly enables 3D scene editing, where the objects observed in an input view can be rendered with novel poses, shapes and/or appearances, effectively creating a novel scene. We provide a example of the scene editing capability in Fig.~\ref{fig:scene_editing} and refer to the supplementary document for further demonstrations.

\begin{comment}
\begin{figure}
\begin{center}
\includegraphics[width=.99\columnwidth]{fig/val_nusc2.png}
\end{center}
\caption{
Qualitative comparison on NuScenes: Baselines trained on image pairs.
}
\label{fig:nusc_results2}
\end{figure}
\end{comment}

\section{Conclusions}

In this work, we proposed a new approach for learning neural 3D object representations that in contrast to the majority of existing works exploits exclusively single views of object instances during training, without leveraging other 3D object shape priors such as CAD models or resorting to curated datasets. 
To address this challenging training setting, our method leverages machine-generated labels, namely 3D object detection and panoptic segmentation, to learn a normalized object-centric representation, which is pose independent and factorizes into a shape and an appearance component. These two components are decoded into an implicit radiance field representation for the object, which can then be rendered into novel target views.

We show that our approach generalizes well to unseen objects, even across different datasets of real-world street scenes. 

\myparagraph{Societal impact and limitations} Our work helps to further investigate the possibilities of leveraging real-world, large-scale data for building representations needed in future AR/VR applications. As for the limitations, our work requires significant computational efforts for producing renderings of novel views, akin to related works from neural representation learning. Further, we will investigate AutoRF's applicability to more articulated object categories.

%%%%%%%%% REFERENCES
{
    % \clearpage
    \small
    \bibliographystyle{ieee_fullname}
    \bibliography{macros,main}
}

% --- supplementary material
% \input{sec/X_supplementary}

% --- uncomment this to read the instructions
% \input{sec/X_instructions}

\end{document}

\newcommand{\myparagraph}[1]{\vspace{3pt}\noindent\textbf{#1.}}

%\title{Learning 3D Object Priors with Implicit Autoencoders}
%\title{Learning 3D Object Priors from Single View Observations}
\title{\vspace{-40pt}AutoRF: Learning 3D Object Radiance Fields from Single View Observations}

\author{
Norman M{\"u}ller$^{1,3}$~~~
Andrea Simonelli$^{2,3}$~~~ 
Lorenzo Porzi$^3$~~~
Samuel Rota Bul\`{o}$^3$~~~\\
Matthias Nie{\ss}ner$^1$~~~
Peter Kontschieder$^3$~~~
\vspace{0.2cm} \\
Technical University of Munich$^1$~~~
University of Trento$^2$~~~
Meta Reality Labs Zurich$^3$
\vspace{0.2cm}
% \and
}

\begin{comment}
\author{Norman M{\"u}ller*\\
Technical University of Munich\\
\and
Andrea Simonelli*\\
University of Trento\\
\and
Lorenzo Porzi\\
Meta Reality Labs\\
\and
Samuel Rota Bul\`{o}\\
Meta Reality Labs\\
\and
Matthias Nie{\ss}ner\\
Technical University of Munich
\and
Peter Kontschieder\\
Meta Reality Labs\\
}
\end{comment}
\maketitle

% --- supplementary material
\appendix

\newpage
\section*{\Large\textbf{Appendix}}
\section{Implementation details}
\paragraph{Encoder}
Our encoder is based on a ResNet34 backbone where we replace all BatchNorm layers with InstanceNorm layers to support batch size of 1.
The first four layers of this architecture are shared while the following two layers are replicated to form separate heads for shape and appearance encoding.
For a $3\times H \times W$ image, input for each encoding head is a feature map of shape $256\times H \mathbin{/}16 \times W \mathbin{/}16$. These feature maps are passed through the individual heads and and adaptive max pooling is applied to obtain shape and appearance codes, each of dimension 128.
We rescale the input images to a maximum of 320px in each dimension while preserving the aspect ratio.

\paragraph{Shape decoder}
The shape decoder is a MLP that is made of 5 ResNet blocks with hidden dimension 128. At each layer, we feed the previous feature map and the positional encoding of the query points. In order to match the dimensionality of the positional encoding with the hidden dimensions of the MLP, we apply a single linear layer and aggregate the output with the intermediate feature maps by a simple per-channel mean pooling.

\paragraph{Color decoder}
For decoding the color, we use a similar architecture as for the shape decoder: A MLP of 5 ResNet blocks with hidden dimensionality of 128 and additional linear aggregations for additional input: As for the shape decoder, we aggregate intermediate features with positional encodings by mean pooling. Furthermore, on the third layer we pass in the same way the output of the corresponding layer of the shape encoder. This enables the color decoder to incorporate estimated shape information. On the final two layers, we pass the view direction (encoded as 3-dimensional vector) to account for view-dependent effects.

\paragraph{Volumetric rendering}
For each ray passing through the unit cube in the normalized object space, we compute the intersection segment and uniformly sample 64 points on this segment.  
During training, we randomly sample 1024 rays per input image and fix the rendering resolution to $80\times 120$px.
For the spatial coordinates, we use positional encoding from NeRF with 6 frequencies. 
At test time, we render each sample at a fixed resolution of $64 \times 64$px.

\paragraph{Hyperparameters}
We train at a batch size of 1 and use the Adam optimizer with a learning of $10^-5$. For test-time optimization, we optimize shape, appearance and camera position using the Adam optimizer at learning rates 0.05, 0.02 and 0.02, respectively, for 32 iterations. We notice that a higher learning rate for color enables the AutoRF to focus on mainly adjusting color values while performing only slight modifications on shape and pose. This way, the optimization eschews strong overfitting to the input view and does not deviate too much from the learnt shape and color code manifold.

\section{Additional ablation results}
\subsection{Auto-decoder variant of AutoRF}
In order to better understand the role of the encoder we perform further ablation studies by using AutoRF in an auto-decoder fashion. To do so, we remove the encoder and only optimize the shape and appearance codes. The initialization of the codes is given by the averages computed on the training set. We optimize the auto-decoder version of our method (AutoRF AutoDecoder) for 128 rounds. The results of these ablations can be found in Tab.~\ref{tab:abl_autoRF}. 

\begin{table}[bht]
\resizebox{\columnwidth}{!}{
\begin{tabular}{@{}lcccc@{}}
nuScenes cars       & PSNR ↑    & SSIM ↑    & LPIPS ↓   & FID ↓     \\ 
\toprule
\Ours{} AutoDecoder on test     & 18.77 & 0.485  & 0.231    &    149.74  \\
\Ours{} AutoDecoder on train     & 18.81 & 0.487  & 0.228    &    134.91  \\
\midrule
\Ours{} on test   & 18.94   & 0.491   & 0.223   & 145.10 \\
\Ours{} on train    & \textbf{18.95}    & \textbf{0.493}   & \textbf{0.210}   & \textbf{106.50}  \\
\bottomrule
\end{tabular}}
\caption{Comparison between AutoRF and its auto-decoder variant (AutoRF AutoDecoder) on nuScenes cars.}
\label{tab:abl_autoRF}
\end{table}

We observe that AutoRF AutoDecoder underperforms on all metrics in comparison with AutoRF. This validates the choice of having an encoder inside the architecture. Furthermore, we observe that the speed of convergence in AutoRF is much faster than its auto-decoder variant. This might be related to the fact that the encoder provides, in a single-shot, an already good initial estimate of the codes.

\begin{figure}
\begin{center}
\includegraphics[width=.99\columnwidth]{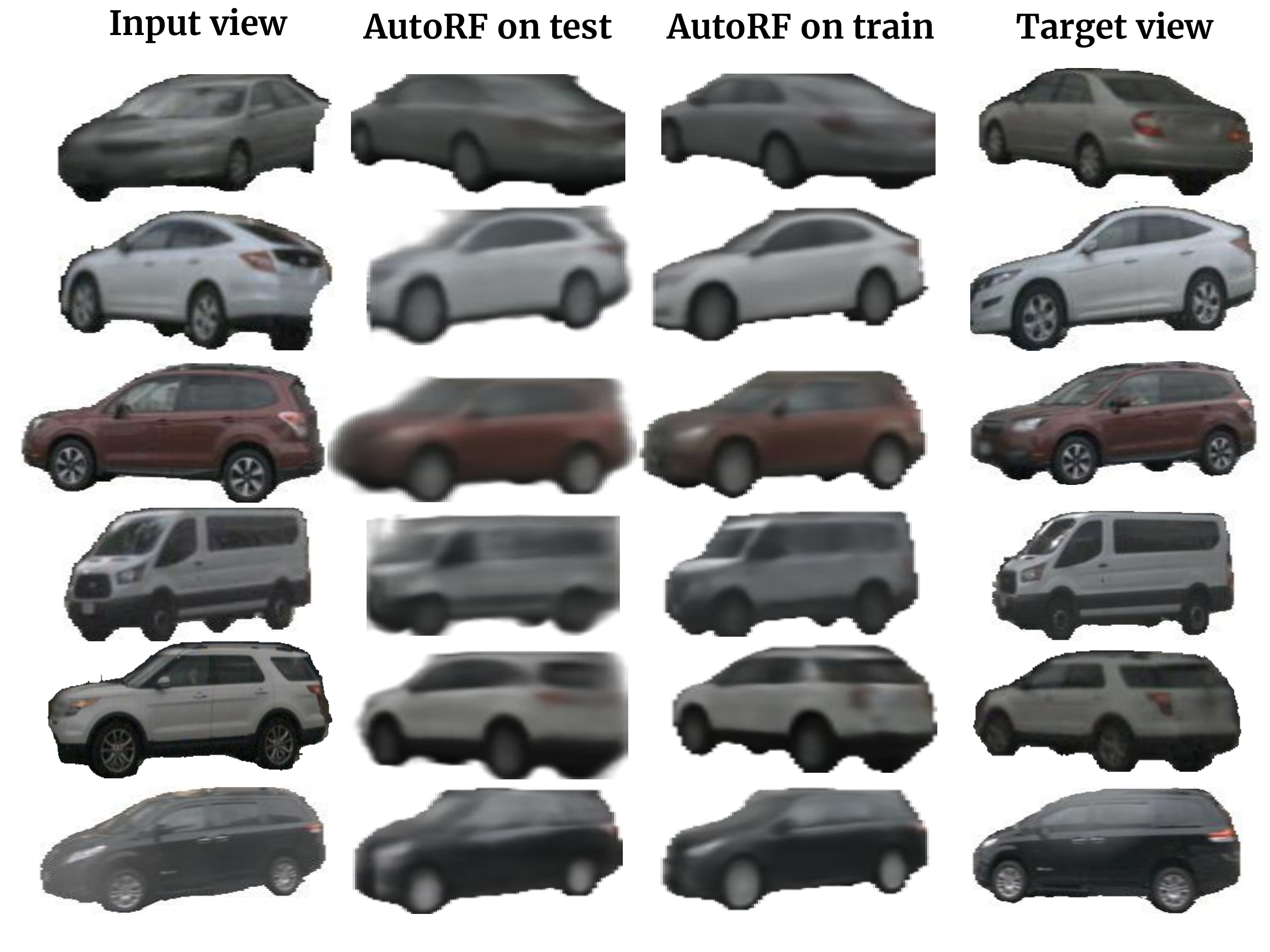}
\end{center}
\vspace{-1pt}
\caption{
Comparison of AutoRF trained on nuScenes test images with machine-generated annotations and AutoRF trained on nuScenes train images with ground-truth annotations.
}
\vspace{-15pt}
\label{fig:train_vs_test}
\end{figure}

\subsection{Evaluation on other categories}
\begin{table}[h]
\includegraphics[width=.99\columnwidth]{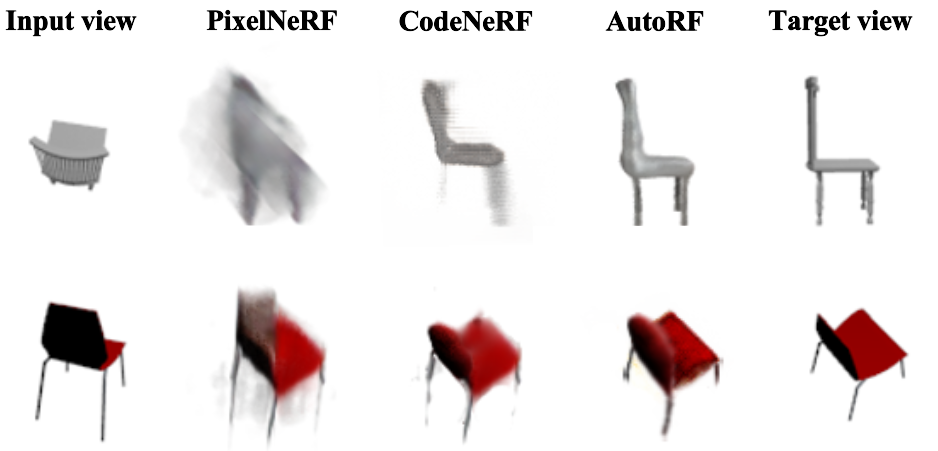}
\label{fig:srn_chairs}
\vspace{-20pt}

\captionof{figure}{Qualitative comparison on SRN-chairs trained on single views.}
\end{table}

\begin{table}[]
\centering
\resizebox{0.85\columnwidth}{!}{
\begin{tabular}{@{}lcccc@{}}
SRN-Chairs       & PSNR ↑    & SSIM ↑    & LPIPS ↓   & FID ↓     \\ 
\toprule
pixelNeRF~[\textcolor{green}{41}]  & 17.73  & 0.726 &  0.218     & 162.9 \\ %\cite{yu2021pixelnerf} 
CodeNeRF~[\textcolor{green}{15}]   &  18.14 &  0.763 &  0.187     & 137.6 \\ %\cite{Jang_2021_ICCV}
\midrule
\Ours{} (no opt.)        & 18.08 & 0.761  & 0.180    & 134.3\\
\Ours{}   & \textbf{18.64}   & \textbf{0.803}   & \textbf{0.148}   &  \textbf{133.2} \\
\bottomrule
\end{tabular}}

\caption{
Evaluation of novel-view synthesis on the SRN-Chairs dataset from~[\textcolor{green}{36}]. % \cite{sitzmann2019scene}.
} % \caption
% \vspace{-10pt}
\label{tab:srn_chairs}
\end{table}

Tab.~\ref{tab:srn_chairs} and Fig.~9, we demonstrate qualitatively and quantitatively that AutoRF performs well also on indoor classes, providing results for SRN-chairs. We follow the same protocol as described in Section~4.2). We notice that pixelNeRF tends to produce more fuzzy radiance fields compared to ours.

%\section{Further analyses}
\subsection{Shape reconstruction}
\begin{figure}[ht!]
\begin{center}
\includegraphics[width=.99\columnwidth]{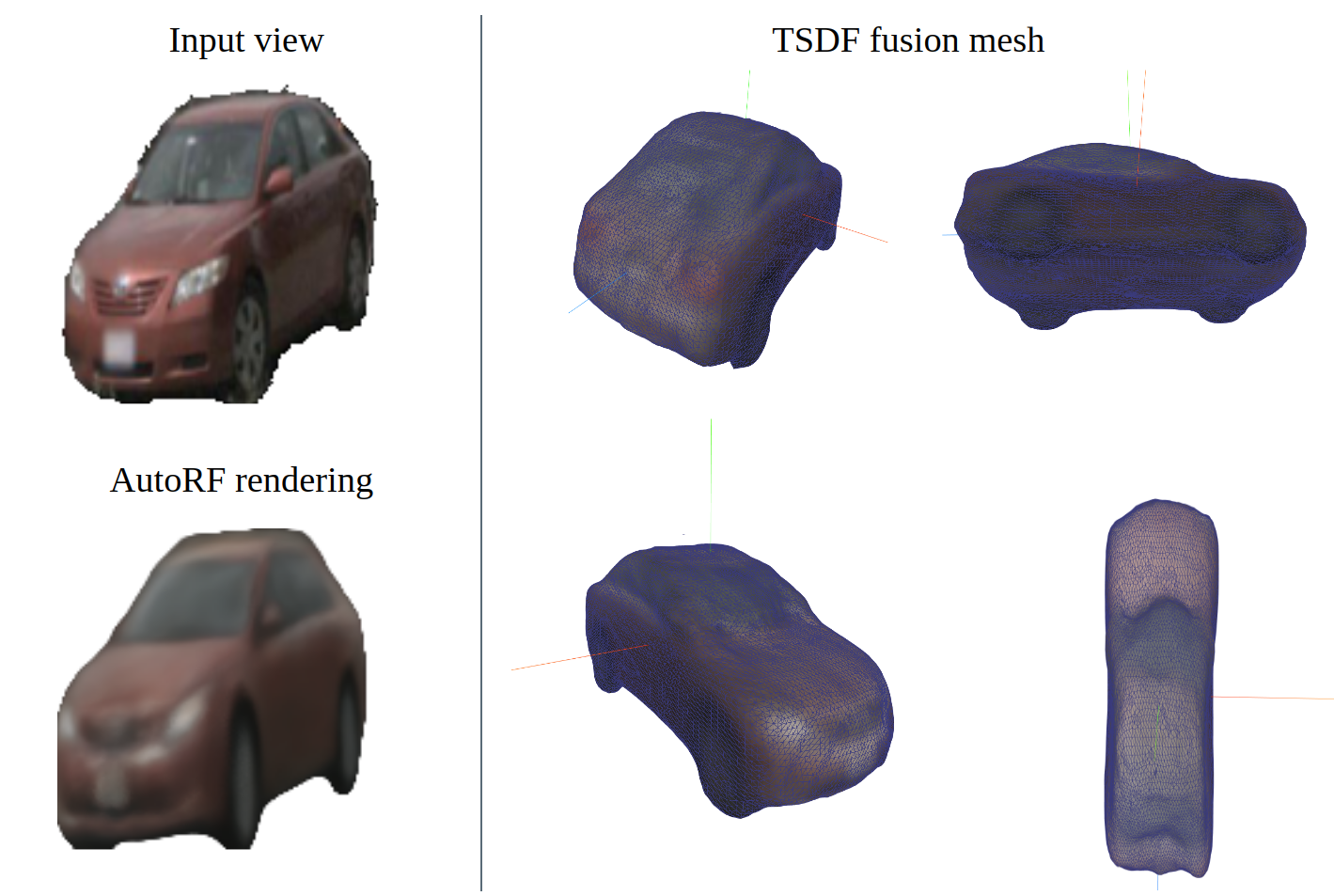}
\end{center}
\vspace{-1pt}
\caption{
Explicit meshing: Given a single input view, we render depth and color from a multiple views using AutoRF and apply TSDF-fusion in order to create a 3D mesh. 
}
\vspace{-15pt}
\label{fig:shape_recon}
\end{figure}

Furthermore, we provide qualitative results of our shape reconstructions in Figure \ref{fig:shape_recon}. For this, we create 40 novel views of the same instance with a camera orbiting around and facing the object center and apply TSDF-fusion on the resulting pairs of depth and color images. We observe consistent depth and color images enabling creating of accurate meshes.

\subsection{Run time}
In this section, we provide further details related to the run time of AutoRF. We observe that the rendering of a $64 \times 64$ image takes approximately 0.23 seconds. Regarding test-time optimization, we observed that the rendering of a $32 \times 32$ image takes approximately 0.11 seconds. Overall the test optimization, which is usually made of 32 steps, takes approximately 3.3 seconds per object.

\section{Further details on the creation of nuScenes novel view data}
We train and evaluate on nuScenes, a data set consisting of 168k training images, 36 validation images, and 36k test images. After filtering for daytime scenes with dry weather, we run the pre-trained 2D panoptic segmentation model [\textcolor{green}{32}] to obtain segmentation masks for all remaining images. 

For the training and validation data, we match provided 3D bounding box annotations with the instance masks resulting from the panoptic segmentation. For the test data, we first run the 3D monocular detection model from [\textcolor{green}{35}], 
filter for detections with a score above 0.7, and match the resulting 3D annotations.
We classify each pixel into instance foreground, background, or unknown region based on their semantic mask. As we do not leverage depth information, we rely on a simple heuristic: Semantic classes that cannot occlude any foreground instance (street, sky, sidewalk, manhole, crosswalks) are considered background while others are considered "unknown" if the do not belong to the object in focus. For those pixels, we do not compute any loss and exclude them during optimization. 
For the generation of the validation set, we leverage the tracking information provided in the ground-truth annotations of nuScenes in order to create image pairs of the same instance at different time steps. We then filter for sufficiently visible instances: we only consider instances where the segmentation mask occupies at least 60\% of the instance 2D bounding box, the instances are no further than 40m distant to the input camera and the overall resolution has to be at least 40px. Based on the remaining images, we randomly select 10k pairs for novel view evaluation.

\section{Analysis of data quality}
In this section, we further discuss the impact of having machine-generated annotations as opposed to manually annotated ones. An initial discussion with quantitative evaluations can be found in the main paper in Sec.~\textcolor{red}{4.3} and Tab.~\textcolor{red}{4}. Here we provide qualitative results in Fig.~\ref{fig:train_vs_test}, where it can be seen that results on test (second column) experience a slight increase of blurriness with respect to the ones on train (third column). It is important to note that, despite the limited decrease in sharpness, AutoRF is able to reliably synthesize novel views on never-seen objects present in the validation data. This is clear from the results shown in e.g. the first row, where AutoRF can recover a plausible car back having observed only its front.

\section{Additional qualitative results }
In this section we provide further qualitative results, aimed at highlighting AutoRF's ability to effectively provide meaningful object representations. 

\subsection{Novel-view synthesis}
The natural output of AutoRF is the rendering of the input image in its original input view. A more interesting output is instead represented by the rendering of the input image from a novel (never-seen) view. Examples of such novel-view synthesis can be found in Fig.~\ref{fig:novel_view_supp} and  Fig.~\ref{fig:novel_metropolis}. It must be noted that in our experiments, AutoRF is focused on learning car representations, so the background, as well as additional objects, are not included into the novel view synthesis.

\subsection{Code interpolation}
AutoRF's explicit disentanglement of the shape and appearance allows to synthesize novel objects by performing a trivial interpolation of the codes. As an example, the appearance code of a red car can be interpolated to the one of a silver van. This results in the synthesis of a novel object smoothly transitioning it's color between red and silver. Interestingly, the same interpolation can be performed on the shape property of the object. Examples of such novel-view synthesis can be found in Fig.~\ref{fig:color_interp_supp}.

\subsection{Scene editing}
\begin{figure}[ht!]
\centering

\resizebox{1.0\columnwidth}{!}{
\begin{tabular}{cc}
    \centering
    \begin{minipage}{2.20\columnwidth}
        \includegraphics[width=.99\columnwidth]{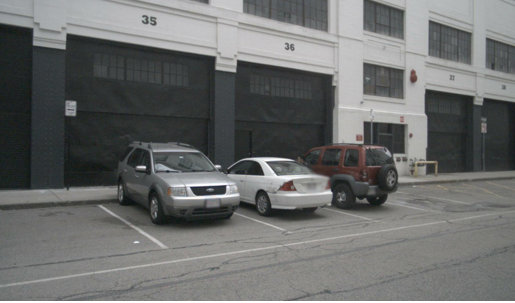}
    \end{minipage} 
    & 
    \begin{minipage}{2.20\columnwidth}
        \includegraphics[width=.99\columnwidth]{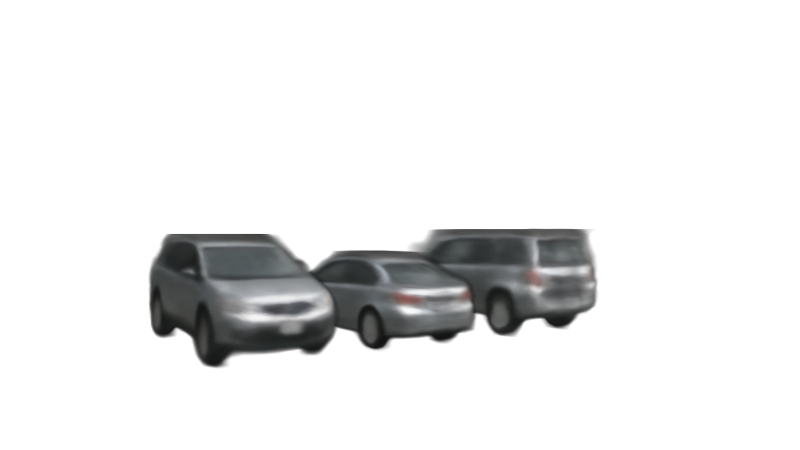}
    \end{minipage} \\
    \begin{minipage}{2.20\columnwidth}
        \includegraphics[width=.99\columnwidth]{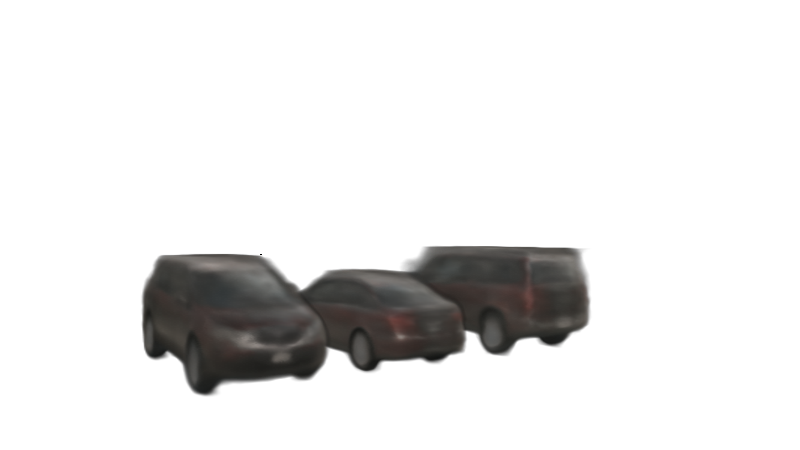}
    \end{minipage}
    &
    \begin{minipage}{2.20\columnwidth}
        \includegraphics[width=.99\columnwidth]{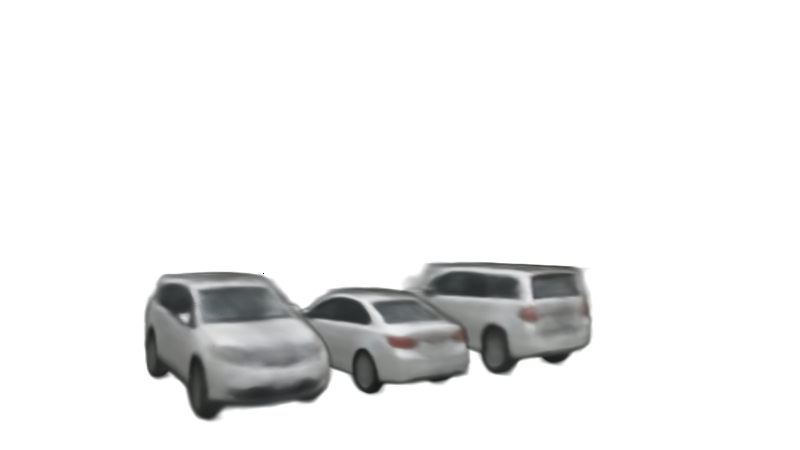}
    \end{minipage}
\end{tabular}}
\caption{
    Color interpolation: starting from the properties of the objects in the input view (top-left), AutoRF is able to modify the properties of each object in the scene.
}
\label{fig:color_interp_supp}

\vspace{2cm}
\resizebox{1.0\columnwidth}{!}{
\begin{tabular}{cc}
    \centering
    \begin{minipage}{2.20\columnwidth}
        \includegraphics[width=.99\columnwidth]{fig/input2.PNG}
    \end{minipage} 
    & 
    \begin{minipage}{2.20\columnwidth}
        \includegraphics[width=.99\columnwidth]{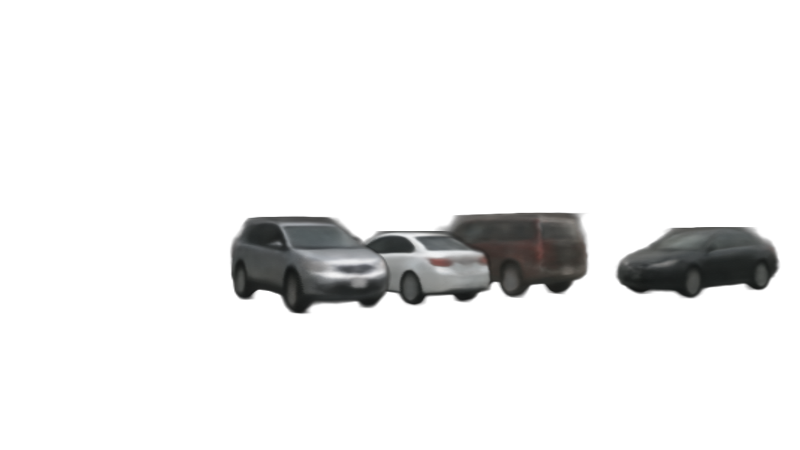}
    \end{minipage} \\
    \begin{minipage}{2.20\columnwidth}
        \includegraphics[width=.99\columnwidth]{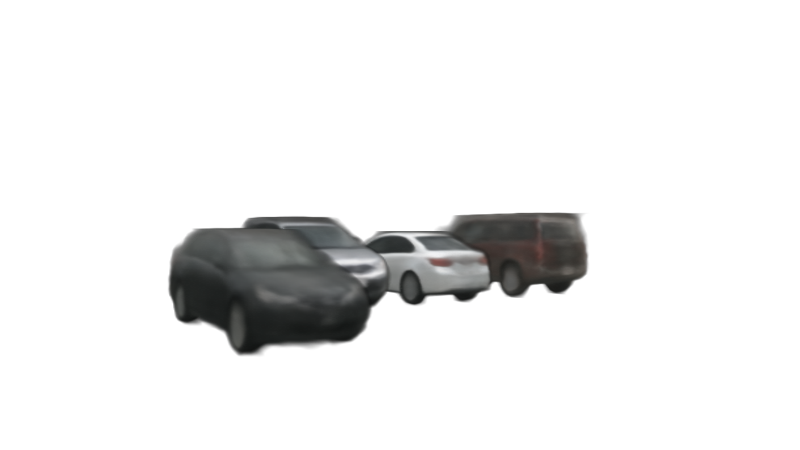}
    \end{minipage}
    &
    \begin{minipage}{2.20\columnwidth}
        \includegraphics[width=.99\columnwidth]{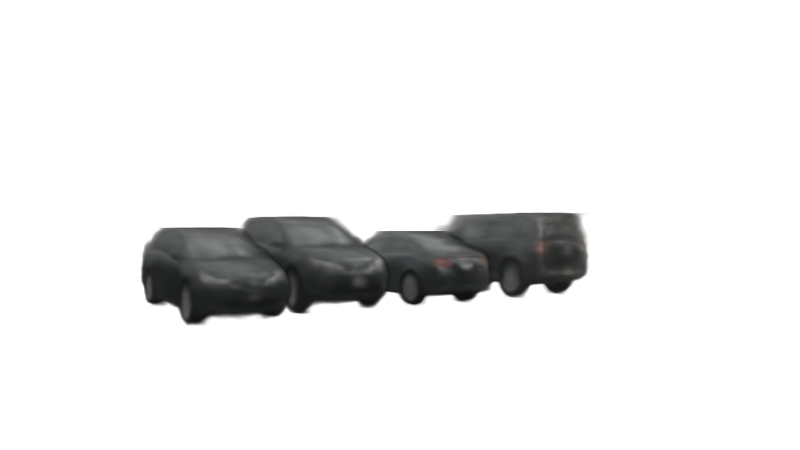}
    \end{minipage}
\end{tabular}}
\caption{
    Scene editing: starting from the objects in the input view (top-left), AutoRF is able to apply arbitrary modifications and also include novel objects. Results on an unseen image of the nuScenes dataset.
}
\label{fig:scene_editing_supp}

\end{figure}

\begin{figure*}
\centering
\resizebox{2.0\columnwidth}{!}{
\begin{tabular}{cc}
    \centering
    \begin{minipage}{2.20\columnwidth}
        \includegraphics[width=.99\columnwidth]{fig/input2.PNG}
    \end{minipage} 
    & 
    \begin{minipage}{2.20\columnwidth}
        \includegraphics[width=.99\columnwidth]{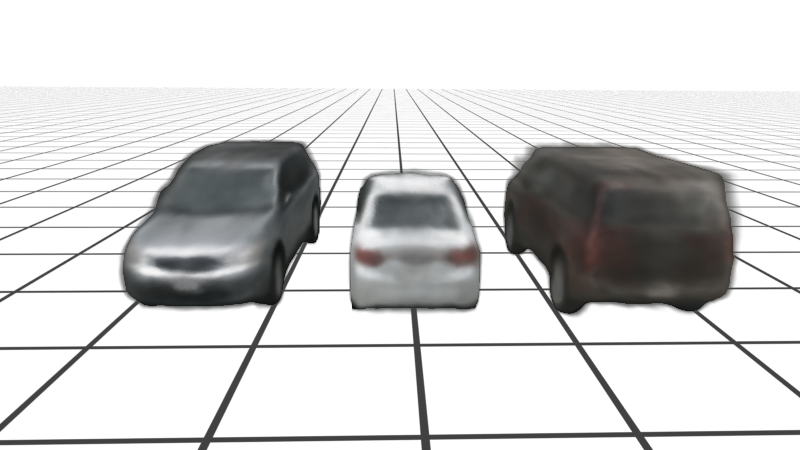}
    \end{minipage} \\
    \begin{minipage}{2.20\columnwidth}
        \includegraphics[width=.99\columnwidth]{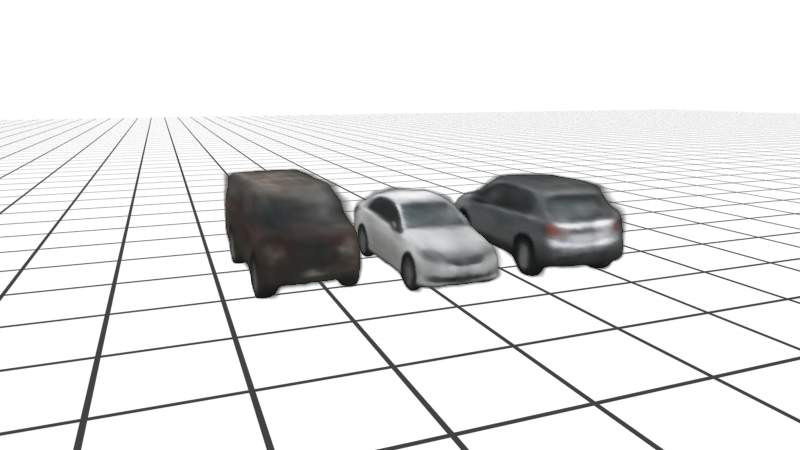}
    \end{minipage}
    &
    \begin{minipage}{2.20\columnwidth}
        \includegraphics[width=.99\columnwidth]{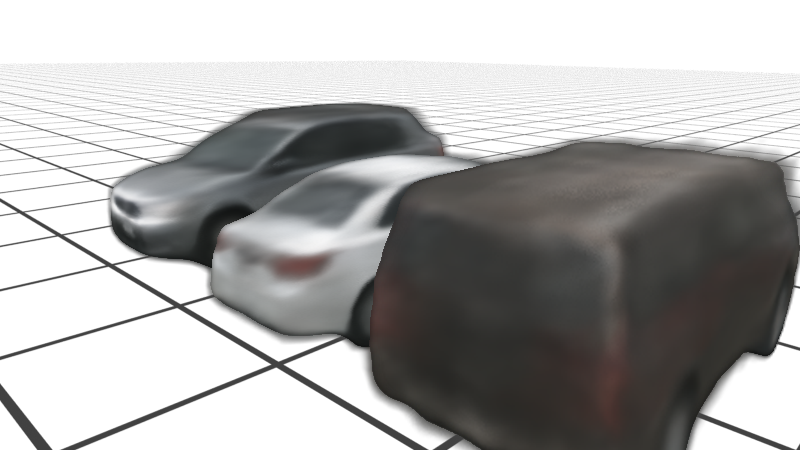}
    \end{minipage}
\end{tabular}}
\caption{
    Novel-view synthesis: by only observing the input view (top-left), AutoRF is able to synthesize the objects in novel views.
}
\label{fig:novel_view_supp}
\end{figure*}

\begin{figure*}

\resizebox{2.0\columnwidth}{!}{
\begin{tabular}{cc}
    \centering
    \begin{minipage}{2.20\columnwidth}
        \includegraphics[width=.99\columnwidth, height=0.5\columnwidth]{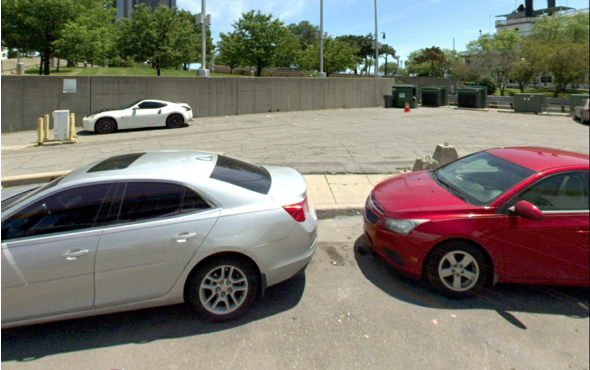}
    \end{minipage} 
    & 
    \begin{minipage}{2.20\columnwidth}
        \includegraphics[width=.99\columnwidth, height=0.7\columnwidth]{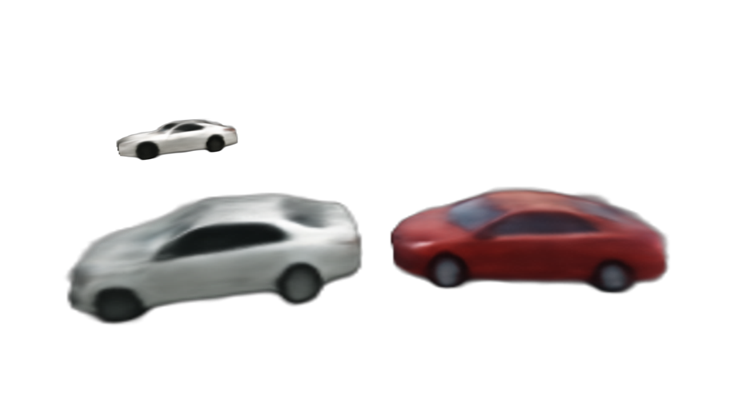}
    \end{minipage} \\
    \begin{minipage}{2.20\columnwidth}
        \includegraphics[width=.99\columnwidth, height=0.7\columnwidth]{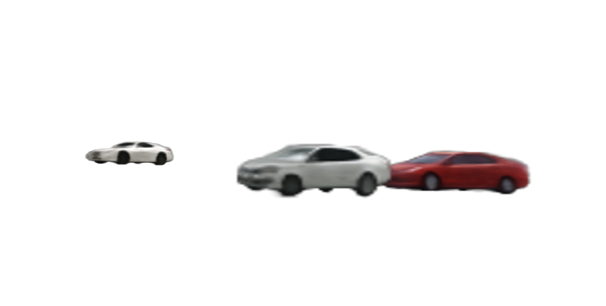}
    \end{minipage}
    &
    \begin{minipage}{2.20\columnwidth}
        \includegraphics[width=.99\columnwidth, height=0.7\columnwidth]{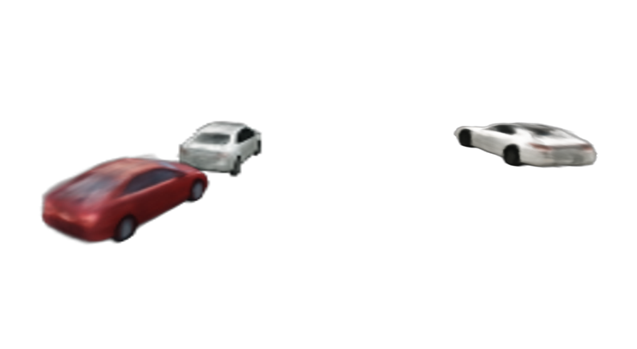}
    \end{minipage}
\end{tabular}}
\caption{
    Novel view synthesis: further results on the unseen Mapillary Metropolis dataset. The model is solely trained on nuScenes test images with different camera setting.
}
\vspace{-10pt}
\label{fig:novel_metropolis}
\end{figure*}

Novel-view synthesis covers the ability of synthesizing a static scene from novel camera poses. Another interesting ability is to synthesize novel scenes by arbitrarily changing object poses, properties as well as camera poses. Examples of such novel-view synthesis can be found in Fig.~\ref{fig:scene_editing_supp}. 


% --- uncomment this to read the instructions
% \input{sec/X_instructions}
%{
    % \clearpage
%    \small
%    \bibliographystyle{ieee_fullname}
%    \bibliography{macros,main}
%}